# Deformation Monitoring of Tunnel using Phase-based Motion Magnification and Optical Flow

**Kecheng Chen**[*1], **Hiroshi Kogi**[2] **and Kenichi Soga**[1]

[1]Civil and Environmental Engineering Department, UC Berkeley, United States

[2]Civil Engineering Technology Division, Shimizu Corporation, Japan

[*]presenting author (email: kecheng_chen@berkeley.edu)

## 1   INTRODUCTION

During construction, continuous monitoring of underground tunnels can facilitate an in-depth understanding of the ground-tunnel interaction and mitigate potential hazards [1]. Traditional vision-based monitoring can directly capture a large range of motion but cannot separate the tunnel's vibration and deformation mode [2]. Phase-based motion magnification (PMM) is a technique to magnify the motion in target frequency bands [3]. But most research related to PMM has focused exclusively on surface structures [4]. Optical flow (OF) is a method for motion calculation and has a much lower computational cost than Digital Image Correlation (DIC) [5]. This paper proposes a PMM-OF-based monitoring framework to quantify the deformation-induced displacements of the underground tunnel. The framework is used to investigate the behavior of an existing tunnel, RMT2, when the nearby underground station was being constructed [2]. R27, shown in Figure 1, was the instrumented RMT2 ring whose center was directly above the tunneling center line. The prism at the crown is S4P3, and the one at the right axis is S4P4. An automatic total station (ATS) was installed to monitor prisms on the wall, and a Canon DSLR 600D camera was installed to monitor the whole tunnel.

## 2   DEFORMATION MONITORING FRAMEWORK

The framework can be explained as the following. First, high-resolution image sequences are downscaled and converted to grayscale to avoid the high computational overhead. Second, the low-pass Gaussian filter is used to correct the non-uniform illumination in the tunnel scenario. Third, the complex steerable pyramid in PMM decomposes images into magnitude and phase components. With temporal filtering, phase components in long change periods (>12h) are increased by multiplying $1+\alpha$, where $\alpha$ is the magnification factor (usually 15). Fourth, the original magnitude and renewed phase components are combined to reconstruct frames in the deformation mode. Then the 2D Wiener filter is used to smooth the artifacts caused by the large motion. Next, the GPU-accelerated deep-learning dense OF algorithm FlowNet2 is used to estimate the absolute motion of each pixel [5]. Finally, the relative motion among pixels at each tunnel ring is evaluated with the spatiotemporal median filter and scaled based on the scaling factor to generate the deformation map. The scaling factor for each tunnel ring can be calculated assuming the dimension of the prisms is known.

## 3 RESULTS & DISCUSSIONS

The comparison of the convergence estimation between PMM-OF and ATS (benchmark) monitoring is shown in Figure 2, where the two trends can match well. The average convergence estimation error of PMM-OF monitoring is 9.04%, caused by the stacking of camera tilting and motion magnification artifacts. Six points are sampled along R27 to generate the deformation shape through PMM-OF monitoring, as shown in Figure 3(b), which can well match the shape estimated by Alhaddad [2] through the combination of ATS monitoring and Sattar Image Tracking (SIT), as shown in Figure 3(a).

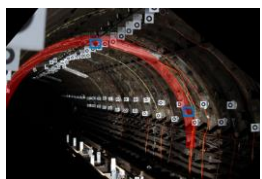

**Figure 1:** Instrumented area (The red area shows R27, and the blue area shows the monitored prisms)

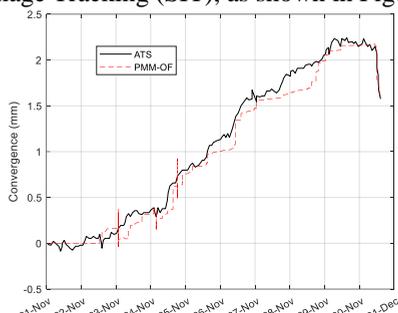

**Figure 2:** Comparison between PMM-OF and ATS monitoring from 11/21/2014 to 12/01/2014 (The convergence between the S4P3 and S4P4)

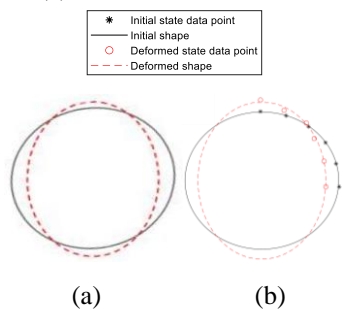

**Figure 3:** Visualization of the tunnel ring deformation on 12/01/2014, where (a) is estimated by Alhaddad [2], and (b) is estimated by PMM-OF

## 4 CONCLUSIONS

This paper presents a PMM-OF-based deformation monitoring framework and estimates the deformation of an existing tunnel disturbed by the nearby construction. The results indicate that the temporal deformation trend and the spatial deformation shape estimated from PMM-OF monitoring can well match that estimated from ATS monitoring.

## REFERENCES

[1] Gue, C. Y., Wilcock, M., Alhaddad, M. M., Elshafie, M. Z. E. B., Soga, K., & Mair, R. J. (2015). The monitoring of an existing cast iron tunnel with distributed fibre optic sensing (DFOS). Journal of Civil Structural Health Monitoring, 5(5), 573-586.

[2] Alhaddad, M. (2016). Photogrammetric monitoring of cast‐iron tunnels and applicabilty of empirical methods for damage assessment. University of Cambridge.

[3] Wadhwa, N., Rubinstein, M., Durand, F., & Freeman, W. T. (2013). Phase-based video motion processing. ACM Transactions on Graphics (TOG), 32(4), 1-10.

[4] Fioriti, V., Roselli, I., Tatì, A., Romano, R., & De Canio, G. (2018). Motion Magnification Analysis for structural monitoring of ancient constructions. Measurement, 129, 375-380.

[5] Ilg, E., Mayer, N., Saikia, T., Keuper, M., Dosovitskiy, A., & Brox, T. (2017). Flownet 2.0: Evolution of optical flow estimation with deep networks. In Proceedings of the IEEE conference on computer vision and pattern recognition (pp. 2462-2470).